\documentclass[envcountsame]{llncs}

 \usepackage{paralist}
\usepackage[utf8]{inputenc}
\usepackage[T1]{fontenc}
\usepackage[numbers]{natbib}
\usepackage{amsfonts}
\usepackage{mathtools}
\usepackage{latexsym, amsmath, amssymb}
\usepackage{stmaryrd}
\usepackage{wasysym}
\usepackage{rotating}
\usepackage{pifont}
\usepackage{booktabs}
\usepackage{array}
\usepackage{multirow}
\usepackage{tikz}
\usepackage{color}
\usepackage{xcolor}
\usepackage{enumitem}
\usepackage{xspace}
\usepackage{thm-restate}
\usepackage{bm}

\usepackage{graphicx}
\usepackage{pifont}
\usepackage{misc}
\usepackage{tikz}

\newcommand{\EX}{\ensuremath{{\sf EX}^{\prob}_{\Fmf,\target}}\xspace}
\newcommand{\e}{\ensuremath{e}\xspace} 
\newcommand{\lab}[2]{\ensuremath{\ell_{#2}(#1)}\xspace}
\newcommand{\target}{\ensuremath{t}\xspace} 
\newcommand{\cmark}{\ding{51}}

\definecolor{Black}  {RGB}{0,0,0}
\newcommand{\examples}{\ensuremath{\Emc}\xspace} 
\newcommand{\hypothesisSpace}{\ensuremath{\Lmc}\xspace} 

\pgfdeclarelayer{edgelayer}
\pgfdeclarelayer{nodelayer}
\pgfsetlayers{edgelayer,nodelayer,main}

\tikzstyle{none}=[inner sep=0pt]
\tikzstyle{invisible}=[inner sep=0pt]
\tikzstyle{rn}=[circle,fill=Red,draw=Black,line width=0.8 pt]
\tikzstyle{gn}=[circle,fill=White,draw=Black,line width=0.8 pt]
\tikzstyle{yn}=[circle,fill=Yellow,draw=Black,line width=0.8 pt]
\tikzstyle{simple}=[circle,fill=White,draw=Black]
\tikzstyle{newstyle1}=[circle,fill=Black,draw=Black,line width=0.3 pt,inner sep=0pt]
\tikzstyle{simple2}=[-,dashed,draw=Black]
\tikzstyle{simpledotted}=[-,dotted,draw=Black]
\tikzstyle{simple}=[-,draw=Black,line width=2.000]
\tikzstyle{new}=[-,draw=Black,line width=2.000]
\tikzstyle{arrow}=[-,draw=Black,postaction={decorate},decoration={markings,mark=at position .5 with {\arrow{>}}},line width=2.000]
\tikzstyle{tick}=[-,draw=Black,postaction={decorate},decoration={markings,mark=at position .5 with {\draw (0,-0.1) -- (0,0.1);}},line width=2.000]
\tikzstyle{newstyle2}=[-latex,draw=Black]
\tikzstyle{newstyle3}=[->,dotted,draw=Black]
\tikzstyle{newstyle6}=[->,dotted,draw=Black]

\usepackage[T1]{fontenc}

\newcommand{\prob}{\ensuremath{\Dmc}\xspace}

\newcommand{\poly}{\ensuremath{\mathit{p}}\xspace}
\newcommand{\FLE}{\ensuremath{{\cal F\!LE}}\xspace}
\usepackage{scalerel,stackengine}

\usepackage{scalerel,stackengine}

\title{Learning Description Logic 
Ontologies
}
\subtitle{Five Approaches. Where Do They Stand?}

\author{Ana Ozaki
}

\institute{ University of Bergen\\
              Department of Informatics \\
              {ana.ozaki@uib.no} 
}

\begin{document}


\maketitle

\begin{abstract}
The quest for acquiring a formal representation of the knowledge of a domain of interest has attracted researchers with various backgrounds into a diverse field called ontology learning. We highlight classical machine learning and data mining approaches that have been proposed for (semi-)automating the creation of description logic (DL) ontologies. These are based on association rule mining, formal concept analysis, inductive logic programming, computational learning theory, and neural networks. We provide an overview of each approach and how it has been adapted for dealing with DL ontologies. Finally, we discuss the benefits and limitations of each of them for learning DL ontologies.

\end{abstract}

\section{Introduction}

The quest for acquiring a formal  representation of the knowledge 
of a domain of interest has attracted researchers with various 
backgrounds and both practical and theoretical inquires 
into a  diverse field called \emph{ontology learning}~\cite{Maedche:2001:OLS:630317.630627,lehmann2014perspectives}. 
In this work, 
we focus on approaches for building description logic (DL) ontologies assuming 
that the vocabulary and the language of the ontology to be created are known. 
The main goal is to find how the symbols of the vocabulary 
should be related, using the logical constructs available 
in the ontology language. 
Desirable goals of an ontology learning process include: 
\begin{enumerate} 
\item the creation of ontologies which are 
\emph{interpretable}; expressions 
 should not be overly complex, redundancies should be avoided; 
\item the support for learnability of DL expressions formulated in \emph{rich ontology languages};  
\item \emph{efficient} algorithms for creating ontologies,  requiring  a \emph{small amount of time and training data};
\item  \emph{limited or no human intervention} requirement;
\item the support for learning in \emph{unsupervised} settings;
\item handling of  \emph{inconsistencies and noise}. 
\end{enumerate}
 Other properties such as explainability and trustability may also be relevant 
 for some approaches. 
 Moreover, once the ontology has been created, it needs to be 
 checked, be maintained, and evolve. This means that other 
 reasoning tasks should also be feasible.
 
Nearly 20 years  after the term ``ontology learning'' was coined by Maedche and Staab~\cite{Maedche:2001:OLS:630317.630627}, 
it is not a surprise that no approach could accomplish such ambitious and conflicting goals.
However, different approaches have  addressed some of these goals. 
We highlight five approaches coming from  machine learning and data mining  which have been proposed for (semi-)automating 
the creation of DL ontologies.
 These are  based on association rule mining (ARM)~\cite{agrawal1993mining}, 
 formal concept analysis (FCA)~\cite{FCA},  inductive logic programming (ILP)~\cite{muggleton1991inductive}, 
  computational learning theory (CLT)~\cite{Valiant}, 
and neural networks (NNs)~\cite{McCulloch:1988:LCI:65669.104377}.

The adaptations of the approaches to the problem of learning DL ontologies 
often come with the same benefits and limitations as the original approach. 
To show this effect, for each of the five approaches, we start by presenting 
the original proposal and then explain how it has been adapted for dealing with 
DL ontologies. 
Before presenting them,  
 we introduce some basic  notions.

\section{Definitions}

Here we  present 
the syntax and semantics of DLs and 
basic definitions useful to formalise learning problems. 

\subsection{Description Logic Ontologies}

We introduce \ALC~\cite{dlhandbook}, a prototypical DL which features basic ingredients 
found in many DL languages.  
Let $\NC$ and $\NR$ be countably infinite and disjoint 
sets of \emph{concept} and \emph{role} names.
An \ALC ontology (or \emph{TBox}) is a finite set 
of expressions of the form $C\sqsubseteq D$, called \emph{concept inclusions} (CIs),
where $C,D$ are \ALC \emph{concept expressions} built according to the 
 grammar rule
\begin{gather*}
C,D::= A \mid \neg C \mid C\sqcap D\mid \exists r.C
\end{gather*}
with $A\in\NC$ and $r\in\NR$. 
An \EL concept expression is an \ALC concept expression without any occurrence 
of the negation symbol ($\neg$). An \EL TBox is a finite set of CIs $C\sqsubseteq D$, 
with $C,D$ being \EL concept expressions. 
 
The semantics of \ALC (and of the \EL fragment) is based on \emph{interpretations}. 
An interpretation \Imc is a pair $(\Delta^\Imc,\cdot^\Imc)$ 
where $\Delta^\Imc$ is a non-empty set, called the \emph{domain of \Imc}, and $\cdot^\Imc$ is a function 
mapping each $A\in\NC$ to a subset $A^\Imc$ of $\Delta^\Imc$
and each $r\in\NR$ to a subset $r^\Imc$ of $\Delta^\Imc\times \Delta^\Imc$. 
The function $\cdot^\Imc$ extends to arbitrary \ALC
concept expressions as follows: 
\begin{align*}
(\neg C)^\Imc &:= {}  \Delta^\Imc\setminus C^\Imc \\
 (C\sqcap D)^\Imc &:= {}  C^\Imc\cap D^\Imc \\
(\exists r.C)^\Imc &:= {}  \{d\in\Delta^\Imc\mid \exists e\in C^\Imc \text{ such that } (d,e)\in r^\Imc\}
\end{align*}
An interpretation \Imc \emph{satisfies} a CI $C\sqsubseteq D$, in symbols $\Imc\models C\sqsubseteq D$, iff 
$C^\Imc\subseteq D^\Imc$. 
It satisfies a TBox \Tmc, in symbols $\Imc\models \Tmc$, iff 
\Imc satisfies all CIs in \Tmc. 
A TBox \Tmc \emph{entails} a CI $\alpha$, in symbols
$\Tmc\models \alpha$, iff all interpretations 
satisfying \Tmc  satisfy $\alpha$.

\subsection{Learning Frameworks}

By \emph{learning} we 
mean the process of acquiring some desired 
kind of   
knowledge
represented in a well-defined and machine-processable form. 
\emph{Examples} are pieces of information that   
characterise such knowledge, given 
as part of the input of a learning process. 
We  formalise these relationships as follows. 

A \emph{learning framework} $\Fmf$ is a triple 
$(\examples, \hypothesisSpace, \mu)$ where  $\examples$ is a set of examples, 
$\hypothesisSpace$ is a set of concept representations\footnote{In the Machine Learning literature,
 a \emph{concept} is often defined as a set of examples and a concept representation is a way of representing such set. 
This  differs from the notion of a concept in the DL literature and 
a formal concept in FCA.}, called \emph{hypothesis space},
and $\mu$ is a function that maps each element of \Lmc to a 
set of (possibly classified) examples in $\examples$. 
If the classification is into $\{1,0\}$, representing positive and negative labels, then $\mu$ simply associates 
 elements $l$ of \Lmc to all examples  labelled with $1$ by $l$. 
Each element of $\hypothesisSpace$ is called a \emph{hypothesis}.
The \emph{target representation} (here simply called \emph{target}) 
is a fixed but arbitrary element of $\hypothesisSpace$,  representing 
the kind of knowledge  that is aimed for in the learning process.  

\begin{example}\label{ex:preli}
To formalise the problem of learning DL ontologies from entailments,
 one can define the learning framework for a given DL $L$ as 
$(\examples, \hypothesisSpace, \mu)$ where 
$\examples$ is the set of all CIs $C\sqsubseteq D$
with $C,D$ being $L$ concept expressions; 
$\hypothesisSpace$ is the set of all $L$ TBoxes; and $\mu$
is a function that maps every $L$ TBox \Tmc to 
the set $\{C\sqsubseteq D\in \examples\mid \Tmc\models C\sqsubseteq D \}$. 
In this case, we consider that  $C\sqsubseteq D$ is labelled with $1$ by \Tmc iff $\Tmc\models C\sqsubseteq D$.
\end{example}

\noindent
In the next five sections, 
we  highlight  
 machine learning and data mining 
approaches which have been proposed for (semi-)automating 
the creation of DL ontologies.
As mentioned, for each approach, we first describe the original motivation and application.   
Then we describe how it has been adapted
 for dealing with DL ontologies.

\section{Association Rule Mining}

\subsection{Original Approach} Association rule mining (ARM) 
is a data mining method frequently used  to 
discover patterns, correlations, or 
causal structures 
in 
transaction databases, relational databases, and other 
information repositories. 
We provide basic notions, as it was initially proposed~\cite{agrawal1993mining}.

\begin{definition}[Association Rule]\label{def:arm}
Given a set $I=\{i_1, i_2,\ldots,i_n\}$  of \emph{items}, and
a set $\mathbb{D} = \{t_1, t_2, \ldots, t_m\}$ of \emph{transactions} 
(called  \emph{transaction database}) with each $t_i\subseteq I$,
an \emph{association rule} is an expression of the form
$A\Rightarrow B$
where $A,B$ are sets of items.
\end{definition}
The task of mining rules  is divided into two parts: (i) mining sets of 
items which are frequent in the database, and, (ii) generating association rules 
based on frequent sets of items. To measure the frequency of a set $X$ of items in a transaction database $\mathbb{D}$, 
one uses a measure  called \emph{support},  defined 
as:
$${\sf supp}_\mathbb{D}(X) =   \frac{|\{ t_i \in \mathbb{D} : X\subseteq t_i\}|}{|\mathbb{D}|}$$
If a set $X$ of items has support larger than a given threshold then it is used in 
the search of association rules, which have the form $A \Rightarrow B$, with $X = A\cup B$. 
To decide whether an implication $A \Rightarrow B$ should be 
in the output of a solution to the problem, 
a confidence measure is used. 
The confidence of an association rule $A \Rightarrow B$ w.r.t. a transaction database $\mathbb{D}$ is defined as:
$${\sf conf}_\mathbb{D}(A \Rightarrow B) =   \frac{{\sf supp}_\mathbb{D}(A\cup B )}{{\sf supp}_\mathbb{D}(A)}$$
Essentially, support   measures   statistical significance, while  
 confidence   measures  the `strength' of a rule \cite{agrawal1993mining}.  
 
We parameterize the ARM learning framework $\Fmf_{\sf ARM}$ with the confidence threshold
$\delta\in  [0,1]\subset\mathbb{R}$.
$\Fmf_{\sf ARM}(\delta)$ is 
$(\examples,\Lmc,\mu)$ where $\examples$ is the set of all database transactions $\mathbb{D}$; 
\Lmc is the set of all sets $\Smc$
of association rules; 
and 
\[\mu(\Smc)=\{\mathbb{D}\in \examples\mid \forall \alpha\in \Smc\text{ we have that } {\sf conf}_\mathbb{D}(\alpha)\geq \delta\}.\]

\noindent\textbf{ARM Problem:}
Given $\delta$ and $\mathbb{D}$, let  $\Fmf_{\sf ARM}(\delta)$ be $(\examples,\Lmc,\mu)$.
Find $\Smc\in\Lmc$ with  $\mathbb{D}\in\mu(\Smc)$. 

\smallskip

 \begin{example}\label{ex:arm}
 Consider the transaction database  in Table~\ref{tab:tdb}.
 It contains $5$ transactions. For example, 
 the first transaction, $t_1$, has ${\sf Product 1},{\sf Product 3}$ and 
 ${\sf Product 4}$ (first row of Table~\ref{tab:tdb}).
Assume that the support and confidence thresholds are resp. $60\%$ and $70\%$. 
 ARM  gives 
the rules: $\{{\sf Product 3},{\sf Product 4}\} \Rightarrow \{{\sf Product 1}\}$ (conf. $75\%$) and 
$\{{\sf Product 2}\} \Rightarrow \{{\sf Product 1}\}$ (conf. $100\%$), among others. 
 \end{example}

\begin{table}[t]
\caption{Transaction Database  }
\label{tab:tdb}
\centering
\begin{tabular}{|c||c|c|c|c|}
\hline
ID &   Product 1 & Product 2 & Product 3 &  Product 4 \\ 
\hline 
\hline
1       &  \cmark  &     & \cmark   & \cmark  \\ 
 \hline
2 		     &    &     &  \cmark  & \cmark \\ 
\hline
3             &   \cmark  &   \cmark  &   \cmark & \cmark \\ 
\hline
4 &       \cmark    &  \cmark  &    \cmark    &\\ 
\hline
5            &    \cmark	& \cmark  & \cmark   &  \cmark \\ 
\hline 
\end{tabular} 
\end{table}

 \subsection{Building DL ontologies}
 An immediate way of adapting the ARM approach to deal with DL ontologies 
 is to make the correspondence between a \emph{finite}  interpretation
 and a transaction database.  
 Assume \Imc is a finite interpretation then the notions 
 of support and confidence can be adapted to: 
 \begin{equation*}
 {\sf supp}_\Imc(C) =   \frac{| C^\Imc|}{|\Delta^\Imc|}
\quad\quad \ {\sf conf}_\Imc(C \sqsubseteq D) =   \frac{{\sf supp}_\Imc(C\sqcap D )}{{\sf supp}_\Imc(C)}
\end{equation*}

 The problem of giving  logical meaning to  association rules  is that 
  it may happen that $C\sqsubseteq D$ and $D\sqsubseteq E$ 
  have confidence values above a certain threshold
  while $C\sqsubseteq E$ \emph{does not} have 
   a confidence value  that is above the threshold,
  even though 
  it is a logical consequence of the first two CIs~\cite{BorDi11}. 
  This problem also occurs in the original ARM approach if 
  the association rules are interpreted as 
  Horn rules  in propositional logic. 
  To see this effect, consider Example~\ref{ex:arm}. We have that both 
  $\{{\sf Product 3},{\sf Product 4}\} \Rightarrow \{{\sf Product 1}\}$  and
  $\{{\sf Product 1}\} \Rightarrow \{{\sf Product 2}\}$ have 
  confidence $75\%$ but the confidence of $\{{\sf Product 3},{\sf Product 4}\} \Rightarrow \{{\sf Product 2}\}$
  is only $50\%$. 
  Another difficulty in this adaptation for dealing with DLs is that 
  the number of CIs 
  with confidence value above a given threshold 
  may be infinite (consider e.g. \EL CIs in an interpretation with a directed cycle)
  and a finite set which implies such CIs may not exist.

The learning framework here is parameterized with a DL $L$
and a confidence threshold $\delta\in [0,1]\subset \mathbb{R}$. Then, 
$\Fmf^{\sf DL}_{\sf ARM}(L,\delta)$ is $(\examples,\Lmc,\mu)$ with $\examples$ the set of all finite  interpretations \Imc; 
\Lmc  the set of all $L$ TBoxes 
\Tmc;
and 
\[\mu(\Tmc)=\{\Imc\in\examples\mid \forall \alpha\in\Tmc\text{ we have that } {\sf conf}_\Imc(\alpha)\geq \delta\}.
\]

 \noindent\textbf{ARM+DL Problem:}
Given  \Imc,  $L$, and  $\delta$, let $\Fmf^{\sf DL}_{\sf ARM}(L,\delta)$ be $(\examples,\Lmc,\mu)$.
Find $\Tmc\in\Lmc$ with $\Imc\in\mu(\Tmc)$.

\medskip

 ARM is an effective approach  for extracting CIs 
 with concept expressions of fixed length from RDF datasets. 
Using this technique, e.g., 
${\sf DeputyDirector} \sqsubseteq {\sf CivilServicePost}$ 
and ${\sf MinisterialDepartment} \sqsubseteq {\sf Department}$
were   extracted from   \url{data.gov.uk}~\cite{volker2011statistical,DBLP:conf/otm/FleischhackerVS12,DBLP:journals/ws/VolkerFS15}
(see also~\cite{DBLP:conf/semweb/SazonauS17} for expressive DLs with fixed length).
 
More recently, ARM has been
applied to mine relational rules
 in knowledge 
graphs~\cite{DBLP:journals/vldb/GalarragaTHS15}.
This approach, born in the field of data mining, 
is relevant
for the task of 
building DL ontologies,
 as it
can effectively find interesting relationships 
between concept and role names. 
However, 
it lacks support for mining CIs with 
existential quantifiers on the right-hand side~\cite{DBLP:conf/rweb/0001GH18}. 

\section{Formal Concept Analysis}
\subsection{Original Approach}
Formal Concept Analysis (FCA) is a mathematical method of data analysis which
describes the relationship between objects and their attributes 
\cite{FCA} (see also~\cite{DBLP:conf/rweb/GanterRS19} for an introduction to this field). 
In FCA, data is represented by formal contexts describing 
the relationship between finite sets of objects and attributes. 
 The notion of a transaction database~(Definition~\ref{def:arm}) is similar to the notion of a formal context~(Definition~\ref{def:fc}).

 \begin{definition}[Formal Context]\label{def:fc}
 A \emph{formal context} is a  triple $(G,M,I)$, where $G$ is a set of 
objects, $M$ is a set of attributes, and $I \subseteq G\times M$ is a binary 
relation between objects and attributes. 
 \end{definition}
 
  A \emph{formal concept} is  a pair $(A,B)$ consisting of a 
 set $A\subseteq G$ of objects (the `extent') and a set $B\subseteq M$ of attributes (the `intent') 
 such that the extent consists of all objects that share the given 
 attributes, and the intent consists of all attributes shared by the given 
 objects. 
   A formal concept $(A_1,B_1)$ is less or equal to a formal concept $(A_2,B_2)$, 
  written $(A_1,B_1)\leq (A_2,B_2)$ iff $A_1\subseteq A_2$. 
The set of all formal concepts ordered by $\leq$
forms a complete lattice. 

\begin{example}\label{ex:fca1}
$
(\{\heartsuit, \diamondsuit\},\{ \text{Attribute 1, Attribute 2}\})
$
is a formal concept in the formal context shown in Table \ref{tab:fca}. 
\end{example}

\begin{table}[t]
\caption{Formal context } 
\label{tab:fca}
\centering
\begin{tabular}{|c||c|c|c|}
\hline
Objects & Attribute 1 & Attribute 2 & Attribute 3  \\ 
\hline 
\hline
$\square$    & \cmark   &    & \cmark         \\ 
 \hline
$\heartsuit$ & \cmark   & \cmark   &          \\ 
\hline
$\bigcirc$ &     &     & \cmark         \\ 
\hline
$\diamondsuit$ & \cmark    & \cmark    &          \\ 
\hline
$\triangle$  &          &    			& \cmark        \\ 
\hline 
\end{tabular} 
\end{table}

In FCA, dependencies between attributes are 
expressed by \emph{implications}---a notion   similar to the notion of an association rule (Definition~\ref{def:arm}). 
 An implication is an expression of the form $B_1\rightarrow B_2$
 where $B_1,B_2$ are sets of attributes. 
  An implication $B_1\rightarrow B_2$ holds in a formal context $(G,M,I)$ if  every
  object having all attributes in $B_1$ also has all attributes in $B_2$.
  A subset $B \subseteq M$ respects an implication $B_1\rightarrow B_2$ if $B_1\not\subseteq B$ or $B_2\subseteq B$. 
  An implication $i$ \emph{follows} from a set $\Smc$ of implications
  if any subset of $M$ that respects all implications from $\Smc$ also respects $i$. 
In FCA, one is essentially interested in computing the implications that hold
in a formal context. A set \Smc of implications that hold 
in a formal context $\mathbb{K}$ is called 
an \emph{implicational base}  for $\mathbb{K}$ if every implication that holds in $\mathbb{K}$
  follows from \Smc. Moreover, there should be no redundancies in \Smc (i.e., 
  if $i\in\Smc$ then $i$ does not follow from $\Smc\setminus\{i\}$). 
  Implicational bases are not unique. A well-studied kind of 
  implicational base (with additional properties) is 
 called
  \emph{stem} (or \emph{Duquenne-Guigues}) base~\cite{Guigues1986,DBLP:conf/rweb/GanterRS19}.

The learning framework for FCA is
$\Fmf_{\sf FCA}=(\examples,\Lmc,\mu)$ with $\examples$ the set of all formal contexts $\mathbb{K}$; 
\Lmc  the set of all   implicational bases \Smc;
and 
\[
\mu(\Smc)=\{\mathbb{K}\in \examples\mid \Smc\text{ is an implicational base for }\mathbb{K}\}.
\]

\noindent\textbf{FCA Problem:}
Given $\mathbb{K}$, let $\Fmf_{\sf FCA}$ be $(\examples,\Lmc,\mu)$.
Find 
$\Smc\in\Lmc$ with $\mathbb{K}\in\mu(\Smc)$.
 

\subsection{Building DL ontologies}\label{sec:fcadl}
 
Approaches to combine FCA and DL have been addressed by many authors  
\cite{Rudolph04exploringrelational,baader2007completing,DBLP:conf/icfca/BaaderD09,BorDi11,DBLP:journals/corr/abs-2102-10689}. 
 %
A common way of bridging the gap between FCA and DL~\cite{DBLP:phd/de/Distel2011}   is the one that  maps  
a finite interpretation $\Imc = (\Delta^{\Imc},\cdot^{\Imc})$ and a finite 
set $S$ of concept expressions  into formal context 
$(G,M,I)$ in such a way that:
\begin{itemize}
\item each  $d\in\Delta^{\Imc}$ corresponds to an 
object $o$ in $G$; 
\item each concept 
expression $C \in S$ corresponds to an  
attribute $a$ in $M$; and 
\item $d \in C^{\Imc}$ if, and only if, $(o,a)\in I$. 
\end{itemize}

The notion of an implication is mapped to the notion of a CI in a DL.
Just to give an idea, if the formal context represented by Table \ref{tab:fca} is induced by 
a DL interpretation then the CI ${\sf Attribute2} \sqsubseteq {\sf Attribute1}$ 
would be a candidate to be added to the ontology. 
The notion of an implicational base is adapted as follows. 
Let \Imc be a finite interpretation and let $L$ be a DL language with 
symbols taken from a finite vocabulary.
An \emph{implicational base for \Imc and   $L$}~\cite{DBLP:phd/de/Distel2011} 
is a non-redundant set \Tmc
of  CIs formulated in $L$ (for short L-CIs) such that for all $L$-CIs 
\begin{itemize}
\item $\Imc\models C\sqsubseteq D$ if, and only if, $\Tmc\models C\sqsubseteq D$.
\end{itemize}

We  parameterize the learning framework $\Fmf^{\sf DL}_{\sf FCA}$ with a DL $L$. 
Then, $\Fmf^{\sf DL}_{\sf FCA}(L)$ is $(\examples,\Lmc,\mu)$,
where \examples is the set of all finite interpretations \Imc, 
\Lmc is the set of all implicational bases \Tmc for $\Imc\in\examples$
and $L$, and 
\[
\mu(\Tmc){=}\{\Imc\in\examples\mid \Tmc \text{ is an implicational base for }\Imc \text{ and } L \}.
\]

\noindent\textbf{FCA+DL Problem:}
Given  \Imc and  $L$, let $\Fmf^{\sf DL}_{\sf FCA}(L)$ be $(\examples,\Lmc,\mu)$.
Find $\Tmc\in\Lmc$ with $\Imc\in\mu(\Tmc)$.
 
\smallskip

\noindent
Similar 
to the difficulty   described for the DL adaptation of the ARM approach, 
there may be no finite implicational base for a given interpretation and DL. 

\begin{figure}
 \centering
\begin{tikzpicture}
	\begin{pgfonlayer}{nodelayer}
		\node [style=newstyle1] (0) at (0.5, -1.5) {};
		\node [style=newstyle1] (1) at (1.25, -1) {};
		\node [style=newstyle1] (2) at (2, -1.75) {};
		\node [style=newstyle1] (3) at (2.75, -1) {};
		\node [style=invisible] (4) at (0.5, -1.75) {$e_1$};
		\node [style=invisible] (5) at (1.25, -1.25) {$e_2$};
		\node [style=invisible] (6) at (2, -2) {$e_4$};
		\node [style=invisible] (7) at (2, -1.5) {$B$};
		\node [style=invisible] (8) at (1.25, -0.75) {$A$};
		\node [style=invisible] (9) at (2.85, -0.65) {$A,B$};
		\node [style=invisible] (10) at (2.75, -1.25) {$e_3$};
		\node [style=invisible] (11) at (2, -0.75) {$r$};
		\node [style=invisible] (12) at (3.2, -0.95) {$r$};
	\end{pgfonlayer}
	\begin{pgfonlayer}{edgelayer}
		\draw [style=newstyle2, in=45, out=-30, loop] (3) to ();
		\draw [style=newstyle2] (1) to (3);
	\end{pgfonlayer}
\end{tikzpicture}
 \caption{$\{A\sqsubseteq \exists r.(A\sqcap B)\}$ is a  base for $\EL_{\sf rhs}$~\cite{klarmanontology}.}
 \label{fig:fca}
\end{figure}
\begin{example}
Consider the interpretation  in Figure~\ref{fig:fca}.
An implicational base for $\EL_{\sf rhs}$---the \EL fragment that allows only conjunctions of concept names 
on the left-side of  CIs---is $\{A\sqsubseteq \exists r.(A\sqcap B)\}$~\cite{klarmanontology}.
However, if we remove $e_3$ from the extension of $A,B$ then, for all $n\in\mathbb{N}$,
the CI $A\sqsubseteq \exists r^n.\top$ holds and there is no $\EL_{\sf rhs}$ finite base 
that can entail all such CIs. 
More expressive languages can be useful for the computation of finite bases. 
It is known that, for \EL with greatest fixpoints semantics, a finite implicational base  
always exists~\cite{DBLP:phd/de/Distel2011}.  

\end{example}

Classical FCA and ARM assume that all the information about the individuals is 
known and can be represented in a finite way. 
A `\cmark' 
in a table representing a formal context means that 
the attribute holds for the corresponding object and the absence means that   
the attribute does not hold. In contrast, DL makes the `open-world'  
assumption, and so, the absence of information indicates a 
lack of knowledge, instead of negation. To deal 
 with the lack of knowledge, the authors of 
  \cite{baader2007completing} introduce the notion of a partial context, in 
  which affirmative and negative information about individuals 
  is given as input and an expert is required to decide whether a given 
   concept inclusion 
   should hold or not.  
   
The need for a finite representation of objects and their attributes 
hinders the creation of 
 concept inclusions expressing, for instance, 
 that `every human has a parent that is a human', in symbols
 \[
 {\sf Human}\sqsubseteq \exists {\sf hasParent}.{\sf Human}
 \]
  or 
 `every natural number has a successor that is a natural number', 
 where elements of a model capturing the meaning of the relation are  linked by an infinite 
 chain.
This limitation is shared by all approaches which mine CIs from data, 
including ARM, but in FCA this difficulty is more evident as it  requires 
100\% of confidence.
This problem 
can be avoided by
allowing the system to interact  with an expert who can  assert domain knowledge that
cannot be conveyed from the finite interpretation given as input~\cite{Rudolph04exploringrelational}.

\section{Inductive Logic Programming}
\subsection{Original Approach}
ILP is an area between logic programming 
and machine learning~\cite{muggleton1991inductive}.
In the general setting of ILP, we are given 
a logical formulation of background knowledge and some 
examples classified into positive and negative \cite{muggleton1991inductive}. 
The background knowledge 
is often formulated with a \emph{logic program}---a 
non-propositional version 
 of Horn clauses where all variables in a clause are 
 universally quantified 
 within the scope of the entire clause. 
The goal is to extend the background knowledge $\Bmc$ with a hypothesis $\Hmc$  
in such a way that 
all examples in the set  of positive examples can be 
deduced from the modified background knowledge
and none of the elements of the set  of negative examples can be deduced from it. 

We  introduce the syntax of function-free first-order Horn clauses. 
A term $t$ is either a variable or 
a constant. 
An \emph{atom} is an expression of the form 
$P(\vec{t})$ with $P$ a predicate and $\vec{t}$ a list of terms $t_1,\ldots,t_a$ 
where $a$ is the arity of $P$. An atom is \emph{ground} if all terms 
occurring in it are constants. 
A \emph{literal} is an atom $\alpha$ or its negation $\neg\alpha$.
A first-order  clause is a universally quantified disjunction of literals. 
It is called \emph{Horn} if it has at most one positive literal.
A \emph{Horn expression} is a set of (first-order) Horn clauses.
A \emph{classified example} in this setting is a pair $(e,\lab{e}{})$ 
where $e$ is a ground atom 
and $\lab{e}{}$ (the label of $e$)
is $1$ if $e$ is a positive example or 
$0$ if it is negative.

\begin{definition}[Correct Hypothesis]\label{def:ilp}
Let \Bmc be a Horn expression and   $S$   a set  of pairs $(e,\lab{e}{})$
with $e$ a ground atom and $\lab{e}{}\in\{1,0\}$. 
A Horn expression $\Hmc$ is a \emph{correct hypothesis for \Bmc and $S$} if  
\[ \forall (e,1)\in S, \ \Bmc\cup \Hmc \models e \text{ and } \forall (e,0)\in S, \ \Bmc\cup \Hmc \not\models e.\]
\end{definition}

\begin{example}
Suppose that we are given as input the background knowledge $\Bmc$
\footnote{We use the equivalent representation of  Horn clauses as implications.}
 and a set 
$S$ of classified examples presented in Table \ref{tab:ilp}. 
In this example, one might conjecture a hypothesis $\Hmc$ which states that: 
\[
\forall xy( {\sf isExpert}(x,y)\wedge {\sf Domain}(y)\rightarrow {\sf DomainExpert}(x)).
\] 
\end{example}
\begin{table}[t]
\caption{Background knowledge and classified examples }
\label{tab:ilp}
\centering
\begin{tabular}{|c|}
\hline
Background Knowledge  \\ 
\hline 
$\forall x({\sf MedicalDomain}(x)\rightarrow {\sf Domain}(x))$ \\ 
${\sf Person}({\sf John})$, ${\sf MedicalDomain}({\sf Allergy})$ \\ 
${\sf isExpert}({\sf John}$, ${\sf Allergy})$  \\ 
\hline 
\hline 
 Classified Examples \\
\hline 
$({\sf DomainExpert}({\sf John}),1)$\\
$({\sf DomainExpert}({\sf Allergy}),0)$\\
\hline 
\end{tabular} 
\end{table}

This form of inference is not sound in the logical sense since $\Hmc$ does not 
necessarily follow from $\Bmc$ and $S$. Another  hypothesis considered as correct by this approach
 would be  
 \[
 \forall x({\sf Person}(x)\rightarrow {\sf DomainExpert}(x)),
 \] 
even though one could easily think of an interpretation with a person not being a domain expert.  
One could also create a situation in which there are infinitely many hypotheses suitable 
to explain the positive and negative examples. For this reason, 
it is often required a non-logical constraint to justify the choice of a particular 
hypothesis \cite{muggleton1991inductive}. 
A common principle  is the Occam's razor principle which says that
the simplest hypothesis is the most likely to be correct
(simplicity can be understood in various ways, a naive way is to consider 
the length of the Horn expression as a string).

We parameterize the learning framework for ILP with the background knowledge \Bmc, 
given as part of the input of the problem. 
We then have that 
$\Fmf_{\sf ILP}(\Bmc)$ is the learning framework $(\examples,\Lmc,\mu)$ with $\examples$ the set of all
ground atoms;
$\Lmc$  the set of all Horn expressions $\Hmc$;
and 
\[
\mu(\Hmc)=\{e \in\examples\mid \Bmc\cup\Hmc\models e\}.\] 

Classified examples help to distinguish 
a target unknown logical theory formulated as a Horn expression
from other Horn expressions
 in the hypothesis space. 
 In the learning framework $\Fmf_{\sf ILP}(\Bmc)$, positive examples for
 a Horn expression $\Hmc$
 are those entailed by the union of $\Hmc$
 and the background theory \Bmc. 
 
\medskip
 
\noindent\textbf{ILP Problem:}
Given \Bmc and $S$ (as in  Definition~\ref{def:ilp}), let $\Fmf_{\sf ILP}(\Bmc)$ be $(\examples,\Lmc,\mu)$.
Find $\Hmc\in\Lmc$ such that $\Hmc$  is a  correct (and simple) hypothesis
for \Bmc and $S$. 
That is, for all $ (e,\lab{e}{})\in S$, $e\in\mu(\Hmc)$ iff $\lab{e}{}=1$.

\smallskip

\subsection{Building  DL Ontologies}
In the DL context, ILP has been applied for learning DL concept 
expressions~\cite{DBLP:conf/ijcai/FunkJLPW19,DBLP:conf/ilp/FanizzidE08,DBLP:journals/apin/IannonePF07,lehmann2009dl,DBLP:journals/ml/LehmannH10,lehmann2010learning}
and for learning logical rules for ontologies~\cite{DBLP:journals/ijswis/Lisi11}. 
We describe here the problem setting for learning DL \emph{concept 
expressions}, which can help the designer to formulate the concept expressions in an ontology. 
As in the classical ILP approach, the learner receives as input 
some background knowledge,   formulated 
as a \emph{knowledge base} $\Kmc = (\Tmc,\Amc)$, 
where \Tmc is a TBox and \Amc is a set of \emph{assertions}, that is, expressions of the form 
$A(a)$, $r(a,b)$ where $A\in\NC$, $r\in\NR$, and $a,b$ are taken from a set \NI of individual names.
Assertions can be seen as ground atoms and \Amc, in DL terms,
is called an \emph{ABox}. 
A set $S$ of 
pairs $(e,\lab{e}{})$ with $e$ an assertion and $\lab{e}{}\in\{1,0\}$ is also given as part of the input.
In the mentioned works, $e$ is of the form 
 ${\sf Target}(a)$, with
  ${\sf Target}$  a concept name in \NC not occurring in \Kmc and $a\in\NI$.  
  
As in the original ILP approach, given $\Kmc=(\Tmc,\Amc)$ and $S$,
a concept expression $C$ (in the chosen DL) is  \emph{correct} for 
\Kmc and $S$ if,
for all $({\sf Target}(a),\lab{{\sf Target}(a)}{})\in S$, we have that 
$(\Tmc\cup\{{\sf Target}\equiv C\},\Amc)\models {\sf Target}(a)$
iff $\lab{{\sf Target}(a)}{}=1$.
\begin{example}\label{ex:ilp}
The background knowledge in Table \ref{tab:ilp} can be  translated 
into  $ (\Tmc,\Amc)$, with 
\[\Tmc = \{{\sf MedicalDomain}\sqsubseteq {\sf Domain}\}\]
 and 
$\Amc$  the set of ground atoms given as background knowledge in Table~\ref{tab:ilp}.
Assuming that the target concept name 
is ${\sf DomainExpert}$ and the set $S$  of classified examples is the one 
in Table \ref{tab:ilp},  correct concept expressions would be $\exists {\sf isExpert}.{\sf Domain}$
and
${\sf Person}$. 
\end{example}

The learning framework and problem statement presented here is for learning \ALC and \EL concept expressions
based on the ILP approach~\cite{lehmann2009ideal,DBLP:journals/ml/LehmannH10}. 
Here the learning framework is parameterized by 
a   knowledge base $(\Tmc,\Amc)$ and a DL $L$.
Then,  $\Fmf^{\sf DL}_{\sf ILP}((\Tmc,\Amc),L)$ is 
$(\examples,\Lmc,\mu)$ where $\examples$ is the set of all
ground atoms;
$\Lmc$  is the set of all $L$ concept expressions $C$ such that 
 ${\sf Target}$ does not occur in it;
and 
\[
\mu(C)=\{e \in\examples\mid (\Tmc\cup\{{\sf Target}\equiv C\},\Amc)\models e\}.\]

\smallskip 
 
\noindent\textbf{ILP+DL Problem:}
Given  $\Kmc$, $L$, and   $S$ (the classified examples),
let  $\Fmf^{\sf DL}_{\sf ILP}(\Kmc,L)$ be $(\examples,\Lmc,\mu)$.
 Find  $C\in\Lmc$ such that $C$ is  correct (and simple) 
 for $\Kmc$ and $S$. 
 That is, for all $ (e,\lab{e}{})\in S$, $e\in\mu(C)$ iff $\lab{e}{}=1$.

\smallskip

\section{Learning Theory}\label{sec:clt}
\subsection{Original Approach}
 We describe two classical learning models in CLT which have been applied 
for learning DL concept expressions and  ontologies. 
 We start with the classical PAC
  learning model and then describe the exact learning model\footnote{The expression ``Probably Approximately Correct'' 
 was coined by Angluin in the paper~\cite{angluinqueries}, 
 where she shows the connection between the two learning models.}.

In the PAC learning model, a learner receives 
classified examples drawn according to a probability distribution
and attempts to create a hypothesis 
that approximates the target.
The aim is to bound 
the probability that a hypothesis constructed by 
the learner misclassifies an example.
This  approach can be applied  
to any learning framework. 
Within this model, one can investigate the complexity 
of learning an abstract target, such as 
a DL concept, an ontology, or the weights   of a NN. 

We now formalise this model.
Let
$\Fmf = (\examples, \hypothesisSpace, \mu)$ be
a learning framework. 
A \emph{probability distribution} $\prob$ over
$\examples$
is a function mapping events in a $\sigma$-algebra $E$ of subsets of $\examples$
to $[0, 1] \subset \mathbb{R}$
such that 
$\prob(\bigcup_{i \in I} X_{i}) = \sum_{i \in I} \prob(X_{i})$ for mutually exclusive $X_i$,
where $I$ is a countable set of indices, $X_{i}\in E$, and
$\prob(\examples) = 1$.
Given a target $\target \in \hypothesisSpace$,
let $\EX$ be the oracle that takes no input, and outputs a 
\emph{classified example}
$(e,\lab{e}{t})$, where $e \in \examples$ is
sampled according to the probability distribution $\prob$, 
  $\lab{e}{t}=1$, if $e \in \mu(\target)$, 
  and $\lab{e}{t}=0$, otherwise. 
An \emph{example query} is a call to the oracle $\EX$.
A 
\emph{sample} generated by $\EX$ is a (multi-)set of indexed classified examples,
independently and identically 
distributed
according to $\prob$, sampled by calling $\EX$.



%
A learning framework
$\Fmf$
is \emph{PAC learnable}
if there is a function $f : (0, 1)^{2} \to \mathbb{N}$
and a deterministic algorithm
such that, for every $\epsilon, \delta \in (0, 1) \subset \mathbb{R}$, every probability distribution $\prob$ on $\examples$,
and every target
$\target \in \hypothesisSpace$,
given a sample 
of size $m \geq f(\epsilon, \delta)$ generated by $\EX$,
the algorithm
always halts and outputs $h \in \hypothesisSpace$ such that
with probability at least $(1 - \delta)$ over the choice of $m$ examples in $\examples$,
we have that $\prob(\mu(h) \oplus \mu(t)) \leq \epsilon$.
If
the number of computation steps used by
the algorithm
is bounded by a polynomial function $\poly(|\target|, |e|, 1/\epsilon, 1/\delta)$,
where
$e$ is the largest example in the sample generated by $\EX$, 
then 
 $\Fmf$
is \emph{PAC learnable in polynomial time}.

\begin{example}
Let 
$\examples=\{\square,\heartsuit,
\bigcirc,\diamondsuit,\triangle\}$ and 
let $\mathcal{D}$ 
be a probability distribution on $\examples$, defined, e.g., by the pairs 
$$(\{\square\}, 0.2),  
(\{\heartsuit\}, 0.1), 
 (\{\bigcirc\}, 0.3), 
 (\{\diamondsuit\}, 0.2),  (\{\triangle\}, 0.2).$$
Assume $h,t\in\Lmc$, 
 $\mu(h)=\{\heartsuit,
\bigcirc\}$, and $\mu(t)=\{\heartsuit, 
\triangle\}$. Then,
the probability $\prob(\mu(h) \oplus \mu(t))$ that $h$ misclassifies 
an example according to \Dmc is $0.5$. 
\end{example} 

\noindent\textbf{PAC Problem:} Given a learning framework 
decide whether it 
is  PAC learnable in polynomial time. 

\smallskip

In the classical PAC approach, the probability 
distribution \Dmc is unknown to the learner. 
The algorithm should provide a probabilistic bound 
for any possible \Dmc.
We now describe the exact learning model.
In this model, a learner tries to 
identify an abstract target known by a teacher, also called 
an \emph{oracle}, by interacting with the teacher~\cite{angluinqueries}. 
The most successful  protocol is based on 
 \emph{membership} and \emph{equivalence} queries. 
As it happens with the PAC learning model, this model can be used 
to formulate learning problems within the context of any kind of    
 learning framework. 

We formalise these notions as follows.
Given a learning framework $\Fmf = (\examples,   \hypothesisSpace, \mu)$, we are interested in the
exact identification of a \emph{target}   concept representation $\target\in\hypothesisSpace$ 
by posing queries to oracles.
Let  ${\sf MQ}_{\Fmf,\target}$ be the oracle that takes as input some $\e \in \examples$ and
returns `yes' if $\e \in \mu(\target)$ and `no' otherwise. 
A membership query is a call to the oracle ${\sf MQ}_{\Fmf,\target}$.
For every $\target \in \hypothesisSpace$, we denote by ${\sf EQ}_{\Fmf,\target}$ the oracle
that takes as input a \emph{hypothesis} concept representation $h \in \hypothesisSpace$
and returns `yes' if $\mu(h) = \mu(\target)$ and a \emph{counterexample} 
$\e \in \mu(h) \oplus \mu(\target)$ otherwise, where $\oplus$ denotes the symmetric
set difference.
There is no assumption regarding which counterexample in $\mu(h) \oplus \mu(\target)$
is chosen by the oracle.  
An equivalence query is a call to the oracle ${\sf EQ}_{\Fmf,\target}$. 
In this model, if examples are interpretations or entailments, 
the notion of `equivalence' coincides with \emph{logical} equivalence. 

A learning framework \Fmf is \emph{exactly learnable}
if there is 
a deterministic algorithm 
such that, for every $\target\in\hypothesisSpace$, 
it eventually halts and outputs some $h\in\hypothesisSpace$ with
$\mu(h) = \mu(\target)$. Such algorithm is allowed to call
 the oracles ${\sf MQ}_{\Fmf,\target}$ and
${\sf EQ}_{\Fmf,\target}$.
If the number of computation steps used by the algorithm 
is bounded by a polynomial $p(|\target|,|\e|)$, where $\target\in \hypothesisSpace$
is the target and $\e \in \examples$ is the largest counterexample seen so far, 
then \Fmf is \emph{exactly learnable in polynomial time}.

\medskip
 
\noindent\textbf{Exact Problem:} Given a learning framework 
decide whether it is exactly learnable in polynomial time. 

\smallskip

In Theorem~\ref{th:transfer}, we recall an interesting
connection between the    exact learning model and the PAC  model 
extended with membership queries. 
If there is a polynomial time algorithm for a learning framework \Fmf 
that is allowed to make membership queries then  $\Fmf$ is \emph{PAC learnable with membership queries in polynomial time}.

\begin{theorem}\cite{angluinqueries}\label{th:transfer}
If a learning framework is exactly learnable in polynomial time 
then it is PAC learnable with membership queries in polynomial time. 
If only equivalence queries are used then it is PAC learnable (without membership queries) in polynomial time.
\end{theorem} 

The converse of Theorem~\ref{th:transfer} does not hold~\cite{Blum:1994:SDM:196751.196815}. 
That is, there is a learning framework that is PAC learnable in polynomial time (even without membership queries) 
but not exactly learnable in polynomial time.

\subsection{Building DL Ontologies}

The PAC learning model has been already applied to learn DL concept expressions 
formulated in DL CLASSIC~\cite{Cohen94learningthe,DBLP:journals/ml/FrazierP96} (see also~\cite{DBLP:conf/dlog/ObiedkovSZ19}). 
The main difficulty in adapting the PAC approach for learning DL ontologies 
is the complexity of this task. 
In the PAC learning model, one is normally interested in \emph{polynomial time} complexity, however, 
many DLs, such as \ALC,  have superpolynomial time complexity 
for the entailment problem
and entailment checks are often important
to combine the information present in the classified examples. 

It has been shown that the \EL fragments 
$\EL_{\sf lhs}$ and 
$\EL_{\sf rhs}$---the \EL fragments that allow only conjunctions of concept names 
on the right-side and on 
the left-side of  CIs, respectively---are polynomial time exactly learnable from 
entailments~\cite{KLOW18,DBLP:conf/aaai/KonevOW16,DBLP:conf/aaai/OzakiPM20,DBLP:conf/rweb/Ozaki20}\footnote{The result for $\EL_{\sf rhs}$ (allowing  conjunctions of concept names on the left-side of CIs) appears  in~\cite[Section~4]{KLOW18}}, 
however, this is not the case for \EL. 
The learning framework is the  one in Example~\ref{ex:preli}
and the problem statement is the same as in the original approach. 
By Theorem~\ref{th:transfer}, the results for $\EL_{\sf lhs}$ and 
$\EL_{\sf rhs}$ are transferable to the PAC learning model 
extended with membership queries. 
The  results show how    changes in the 
ontology language can 
impact the complexity
of searching for a suitable 
 ontology in the hypothesis space.   
The main difficulty of implementing this model is that it is based on oracles, in particular, 
on an equivalence query oracle. 
Fortunately, as already  mentioned,
such equivalence queries can be simulated 
by the sampling oracle of the PAC learning model to achieve 
PAC learnability (Theorem~\ref{th:transfer})\cite{angluinqueries}.

\section{Neural Networks}

\subsection{Original Approach}

NNs are widespread architectures inspired 
by the structure of the brain~\cite{McCulloch:1988:LCI:65669.104377}. 
They may differ from each other not only regarding their weight and activation functions but also structurally, e.g.,   feed-forward NNs are acyclic while
recurrent NNs have cycles. 
One of the simplest models is the one given by Definition~\ref{def:nn}.

\begin{definition}[Neural Network]\label{def:nn} 
An NN
is a triple $(G, \sigma, w)$ where $G=(V,E)$ is a graph, 
with $V$  a set of  nodes, called \emph{neurons}, and 
$E\subseteq V\times V$  a set of (directed) edges; 
$\sigma:\mathbb{R}\rightarrow \mathbb{R}$ is 
the \emph{activation} function;
and $w:E\rightarrow \mathbb{R}$ is 
 the \emph{weight} function. 
\end{definition}

Other parameters that can be part of the definition of an NN are the propagation function 
and  biases.
A widely used propagation function is the weighted sum. The propagation function 
specifies how the outputs of the neurons connected to a neuron $n$ are combined 
to form the input of $n$. Given an input to a neuron, the   activation function 
  maps it to the output of the neuron. 
In symbols, 
the input ${\sf in}(n)$ of a neuron 
$n$ is \[\sum_{m:(m,n)\in E} \sigma({\sf in}(m))\cdot w((m,n)).\] 

The structure of an NN is organized in \emph{layers}, basically, 
an \emph{input}, an \emph{output}, and  (possibly several) \emph{hidden} layers.
The input of an NN is a vector of numbers in $\mathbb{R}$, given as input 
to the neurons in the input layer. The output of the NN
is also a vector of numbers in $\mathbb{R}$, constructed using the 
outputs of the neurons in the output layer. The dimensionality of the input 
and output of an NN varies according to the learning task. 
One can then see an NN as a function mapping an input vector $\vec{x}$ to 
an output vector $\vec{y}$. In symbols, $(G, \sigma, w)(\vec{x})=\vec{y}$.

The main task is to find a weight function that minimizes the \emph{risk} of the NN \Nmc, 
modelled by a function $L_\Dmc(\Nmc)$, with \Dmc a probability distribution on a set of 
pairs $(\vec{x},\vec{y})$ of input/output vectors~\cite{shalev2014understanding}.
The risk of an NN represents how well we expect the NN 
to perform while predicting the classification of unseen examples.

The learning framework can be defined in various ways. 
Here we parameterize it by a graph structure 
and an activation function $\sigma$. We have that $\Fmf_{\sf NN}(G,\sigma)$, with $G=(V,E)$,
is $(\examples,\Lmc,\mu)$ where $\examples$ is a set of pairs $(\vec{x},\vec{y})$
representing input and output vectors of numbers in $\mathbb{R}$ (respectively, and with appropriate dimensionality); 
\Lmc is the set of all weight functions  $w:E\rightarrow \mathbb{R}$; 
and 
\[\mu(w)=\{(\vec{x},\vec{y})\in\examples\mid (G, \sigma, w)(\vec{x})=\vec{y}\}.\] 
One can formulate the NN problem as follows. 
 
\medskip
  
\noindent\textbf{NN Problem:}
Given $G$ and $\sigma$,  
let $\Fmf_{\sf NN}(G,\sigma)$ be $(\examples,\Lmc,\mu)$. Find $w\in\Lmc$ that minimizes the risk 
$L_\Dmc(\Nmc)$
of $\Nmc=(G,\sigma,w)$, where \Dmc is a fixed but arbitrary and unknown probability distribution on \examples.

\medskip

Classified examples for training and validation can be obtained by calling the sampling oracle 
${\sf EX}^\Dmc_{\Fmf,t}$ (recall  ${\sf EX}^\Dmc_{\Fmf,t}$ from Section~\ref{sec:clt}), 
where 
$t\in\Lmc$ is the (unknown) target weight function. 
One of the main challenges of this approach is that finding an optimal weight function 
is computationally hard. Most works apply a heuristic search based on the gradient descent algorithm~\cite{shalev2014understanding}.

\subsection{Building DL Ontologies}

NNs have been applied to learn CIs from sentences expressing definitions, 
called \emph{definitorial sentences}~\cite{Petrucci:2016:OLD:3092960.3092992} (see also~\cite{DBLP:conf/aime/MaD13} for 
more work on definitorial sentences in a DL context, 
and, e.g.~\cite{DBLP:conf/nips/BordesUGWY13,DBLP:journals/corr/YangYHGD14a}, for work on learning assertions based on NNs). 
More specifically, the work on~\cite{Petrucci:2016:OLD:3092960.3092992} is based on   \emph{recurrent} NNs, which are useful to process 
sequential data. The structure of the NN, in this case, takes the form of a grid. 
The authors learn \ALCQ CIs, where \ALCQ is the extension of \ALC with qualified number restrictions.
For example, 
``A car  is a motor vehicle that has $4$ tires and transport people.'' 
  corresponds to 
$${\sf Car}\sqsubseteq {\sf MotorVehicle} \ \sqcap = 4 {\sf has}.{\sf Tires}\ \sqcap \ \exists {\sf transport}.{\sf People}.$$

The main benefits of this approach is that NNs can deal with natural language variability.
The authors  provide an end-to-end solution that does not even require natural 
language processing techniques. 
However, the approach is based on the \emph{syntax} of the sentences, not on their 
semantics, and they cannot capture portions of knowledge across different sentences~\cite{Petrucci:2016:OLD:3092960.3092992}. 
Another difficulty of adapting this approach for learning DL ontologies 
is the lack of datasets available 
for training. 
Such dataset should consist of a large set of pairs of definitorial sentences and their 
corresponding   CI. 
The authors created a synthetic dataset to perform their experiments. 

The learning framework and problem statement for learning DL CIs based on the NN approach~\cite{Petrucci:2016:OLD:3092960.3092992} can be formulated as follows.
The learning framework  for a DL $L$ can be defined as
$\Fmf^{\sf DL}_{\sf NN}(G,\sigma,L)=(\examples, \hypothesisSpace,\mu)$ where
 $\examples$ is a set of  pairs $(\vec{x},\vec{y})$
 with $\vec{x}$  a vector representation of a definitorial sentence and 
 $\vec{y}$  a vector representation of an $L$ CI; and \Lmc and $\mu$ are as   
 in the original NN approach. 
 
\medskip
 
\noindent\textbf{NN+DL Problem:} 
Given $G$, $L$ and $\sigma$,  
let $\Fmf^{\sf DL}_{\sf NN}(G,\sigma,L)$ be $(\examples,\Lmc,\mu)$. Find $w\in\Lmc$ that minimizes the risk 
$L_\Dmc(\Nmc)$
of $\Nmc=(G,\sigma,w)$, where \Dmc is a fixed but arbitrary and unknown probability distribution on \examples.

\section{Where Do They Stand?}

We now discuss the main benefits and limitations of ARM, FCA, ILP, CLT, and NNs for building DL ontologies,
considering the    goals listed in the Introduction. 

\mypar{Interpretability}
refers to the easiness of understanding the learned DL ontology/concept expressions
and obtaining insights about the domain. 
In ARM, the requirement for computing CIs with high support 
often results in highly interpretable CIs  (at the cost of  
fixing the length of concept expressions). 
The FCA approach classically
deals with redundancies, which is often  not considered in ARM approaches. 
However, the CIs generated with this approach 
can 
be 
difficult to interpret~\cite{BorDi11}.
The ILP approach follows  the Occam's razor principle,
which contributes to the generation of interpretable
DL   expressions, although there is no guarantee for the quality of the approximation. 
Such guarantees can be found in CLT,
where the goal is to approximate or exactly 
identify the target. 
However, the focus of these approaches is on  accuracy rather than interpretability. 
Regarding NNs, 
the complex models can deal with high variability in the data but may lose on interpretability.

\mypar{Expressivity} 
refers to 
the expressivity of the DL language supported by the learning process. 
As we have seen, many previous approaches 
 for learning DL ontologies focus on Horn fragments such as \EL~\cite{KLOW18,DBLP:conf/aaai/KonevOW16,DBLP:conf/kr/DuarteKO18,lehmann2009ideal,DBLP:conf/icfca/BaaderD09,BorDi11} 
 (or Horn-like fragments such as \FLE~\cite{Rudolph04exploringrelational}).
 Non-horn fragments have been investigated for learning DL
 ontologies~\cite{volker2011statistical,DBLP:conf/semweb/SazonauS17}
 and concept expressions~\cite{DBLP:conf/ilp/FanizzidE08,DBLP:journals/apin/IannonePF07,DBLP:journals/ml/LehmannH10}
 (fixing the length of concept expressions). 
 As mentioned, $\ALCQ$ CIs can be learned with  NNs~\cite{Petrucci:2016:OLD:3092960.3092992} (see also~\cite{DBLP:journals/kbs/ZhuGPZXQ15}).

\mypar{Efficiency} 
refers to the amount of time and memory consumed by algorithms in order to 
 build a DL ontology (or concept expressions) 
in the context of  a particular approach or a learning model. 
In CLT one can formally establish complexity results for learning problems. 
In ARM the search space is heavily constrained by the support function, 
 which means that usually large portions of the search space can be eliminated 
in this approach. 
The Next-closure algorithm used in FCA is polynomial in the output and has 
polynomial delay, meaning that from the theoretical point of view 
it has interesting properties regarding efficiency. However, 
in practice, there may be difficulties in processing large portions 
of data provenient of knowledge graphs, such as DBpedia~\cite{BorDi11}.

\mypar{Human interactions} may be required 
to  complete 
the information given as input or to validate the 
  knowledge that cannot be represented in a 
finite dataset or in a finite interpretation (recall the 
case of an infinite chain of objects in Subsection~\ref{sec:fcadl}). 
 Since the input is simply a database or an interpretation, the ARM and FCA approaches 
 require  limited or no  human intervention. 
It is worth to point out that 
some DL adaptations of the FCA approach 
 depend on an expert which resembles a membership oracle. 
The difference is that in the exact learning model the 
membership oracle answers with `yes' or `no', whereas 
in FCA the oracle also provides a counterexample 
if the answer is `no'~\cite{Rudolph04exploringrelational}. 
In ILP, examples need to be classified into positive and negative, 
which may require human intervention to classify the examples
before learning takes place. 
The same happens with the CLT models presented. 
The exact learning model  is purely 
based on interactions with an oracle, which 
can be an expert (or even a neural network~\cite{DBLP:conf/icml/WeissGY18}).

\mypar{Unsupervised learning}
is supported by the ARM and FCA approaches, 
as well as some NNs (but not by the DL adaptation we have seen in the literature~\cite{Petrucci:2016:OLD:3092960.3092992}).
As already mentioned, the approaches based on ILP and CLT fall 
in the supervised setting. That is, examples receive some sort of (usually binary) classification.

\mypar{Inconsistencies and noise} are often present 
in the data. 
 The ARM approach  deals with 
them  by 
only requiring that the confidence of the 
CI is above a certain threshold
(instead of requiring that the CI is fully satisfied, as in FCA). 
ILP and CLT classically do not support inconsistencies and noise,
though, the PAC model has an agnostic version in which 
it may not be possible to construct a hypothesis consistent 
with the positive and negative examples (due e.g. to noise in the classification).
NNs can deal very well with data variability, including 
cases in which there are inconsistencies and noise.

\section{Conclusion}
We   discussed  benefits and limitations 
of, namely, ARM, FCA, ILP, CLT, and NNs for DL settings.
Not many authors have applied NNs for learning DL ontologies (when the focus is on building the logical expressions),
even though NNs are widespread in many areas. 
We believe that more works  exploring this approach are yet to come. 
One of the challenges is how to capture the \emph{semantics} of the 
domain. Promising frameworks for capturing the semantics of logical expressions~\cite{DBLP:conf/gcai/Galliani19,DBLP:conf/dlog/PorelloKRTGM19} 
and modelling logical rules~\cite{DBLP:conf/kr/Gutierrez-Basulto18}
 have  been recently proposed.

Finally, we have seen that each approach addresses some of the 
desired properties of an ontology learning process. 
An interesting question is whether they can be combined so as to obtain the best 
of each approach. Indeed, recent works have  proposed ways 
of combining FCA with the exact and PAC learning models~\cite{DBLP:conf/dlog/ObiedkovSZ19}.
Moreover, 
the support and confidence measures from ARM could 
also be applied in FCA for dealing with noisy and incomplete data~\cite{BorDi11}.

\mypar{Acknowledgements}
This work is supported by the Free University of Bozen-Bolzano through the projects PACO and MLEARN.

\bibliographystyle{spmpsci}

\begin{thebibliography}{10}
\providecommand{\url}[1]{{#1}}
\providecommand{\urlprefix}{URL }
\expandafter\ifx\csname urlstyle\endcsname\relax
  \providecommand{\doi}[1]{DOI~\discretionary{}{}{}#1}\else
  \providecommand{\doi}{DOI~\discretionary{}{}{}\begingroup
  \urlstyle{rm}\Url}\fi

\bibitem{agrawal1993mining}
Agrawal, R., Imieli\'{n}ski, T., Swami, A.: Mining association rules between
  sets of items in large databases.
\newblock Special Interest Group on Management Of Data {SIGMOD} \textbf{22}(2),
  207--216 (1993)

\bibitem{angluinqueries}
Angluin, D.: Queries and concept learning.
\newblock Machine Learning \textbf{2}(4), 319--342 (1988)

\bibitem{dlhandbook}
Baader, F., Calvanese, D., McGuinness, D., Nardi, D., Patel-Schneider, P.
  (eds.): The Description Logic Handbook: Theory, Implementation, and
  Applications, second edn.
\newblock Cambridge University Press (2007)

\bibitem{DBLP:conf/icfca/BaaderD09}
Baader, F., Distel, F.: Exploring finite models in the description logic.
\newblock In: ICFCA, pp. 146--161 (2009)

\bibitem{baader2007completing}
Baader, F., Ganter, B., Sertkaya, B., Sattler, U.: Completing description logic
  knowledge bases using formal concept analysis.
\newblock In: {IJCAI}, vol.~7, pp. 230--235 (2007)

\bibitem{Blum:1994:SDM:196751.196815}
Blum, A.L.: Separating distribution-free and mistake-bound learning models over
  the boolean domain.
\newblock SIAM J. Comput. \textbf{23}(5) (1994)

\bibitem{BorDi11}
{Borchmann}, D., {Distel}, F.: Mining of $\mathcal{EL}$-{G}{C}{I}s.
\newblock In: The 11th IEEE International Conference on Data Mining Workshops.
  Vancouver, Canada (2011)

\bibitem{DBLP:conf/nips/BordesUGWY13}
Bordes, A., Usunier, N., Garc{\'{\i}}a{-}Dur{\'{a}}n, A., Weston, J.,
  Yakhnenko, O.: Translating embeddings for modeling multi-relational data.
\newblock In: Advances in Neural Information Processing Systems. NeurIPS, pp.
  2787--2795 (2013)

\bibitem{Cohen94learningthe}
Cohen, W.W., Hirsh, H.: Learning the {C}{L}{A}{S}{S}{I}{C} description logic:
  Theoretical and experimental results.
\newblock In: KR, pp. 121--133 (1994)

\bibitem{DBLP:phd/de/Distel2011}
Distel, F.: Learning description logic knowledge bases from data using methods
  from formal concept analysis.
\newblock Ph.D. thesis, Dresden University of Technology (2011)

\bibitem{DBLP:conf/kr/DuarteKO18}
Duarte, M.R.C., Konev, B., Ozaki, A.: Exactlearner: {A} tool for exact learning
  of {EL} ontologies.
\newblock In: {KR}, pp. 409--414 (2018)

\bibitem{DBLP:conf/ijcai/FunkJLPW19}
Funk, M., Jung, J.C., Lutz, C., Pulcini, H., Wolter, F.: Learning description
  logic concepts: When can positive and negative examples be separated?
\newblock In: {IJCAI}, pp. 1682--1688 (2019)
  
\bibitem{DBLP:conf/ilp/FanizzidE08}
Fanizzi, N., d'Amato, C., Esposito, F.: {DL-FOIL} concept learning in
  description logics.
\newblock In: {ILP}, pp. 107--121 (2008)

\bibitem{DBLP:conf/otm/FleischhackerVS12}
Fleischhacker, D., V{\"o}lker, J., Stuckenschmidt, H.: Mining {RDF} data for
  property axioms.
\newblock In: On the Move to Meaningful Internet Systems: OTM 2012, pp.
  718--735. Springer (2012)

\bibitem{DBLP:journals/ml/FrazierP96}
Frazier, M., Pitt, L.: Classic learning.
\newblock Machine Learning \textbf{25}(2-3), 151--193 (1996)

\bibitem{DBLP:journals/vldb/GalarragaTHS15}
Gal{\'{a}}rraga, L., Teflioudi, C., Hose, K., Suchanek, F.M.: Fast rule mining
  in ontological knowledge bases with {AMIE+}.
\newblock {VLDB} J. \textbf{24}(6), 707--730 (2015)

\bibitem{DBLP:conf/gcai/Galliani19}
Galliani, P., Kutz, O.,  Porello, D., 
                Righetti G., and
                Troquard, N.:  On Knowledge Dependence in Weighted Description Logic
\newblock In: {GCAI}, pp. 68--80 (2019)

\bibitem{DBLP:conf/rweb/GanterRS19}
Ganter, B., Rudolph, S., Stumme, G.: Explaining data with formal concept
  analysis.
\newblock In: {RW}, pp. 153--195 (2019)

\bibitem{FCA}
Ganter, B., Wille, R.: Formal Concept Analysis: Mathematical Foundations.
\newblock Springer (1997)

\bibitem{DBLP:conf/kr/Gutierrez-Basulto18}
Guti{\'{e}}rrez{-}Basulto, V., Schockaert, S.: From knowledge graph embedding
  to ontology embedding? an analysis of the compatibility between vector space
  representations and rules.
\newblock In: {KR}, pp. 379--388 (2018)

\bibitem{DBLP:journals/apin/IannonePF07}
Iannone, L., Palmisano, I., Fanizzi, N.: An algorithm based on counterfactuals
  for concept learning in the semantic web.
\newblock Appl. Intell. \textbf{26}, 139--159 (2007)

\bibitem{klarmanontology}
Klarman, S., Britz, K.: Ontology learning from interpretations in lightweight
  description logics.
\newblock In: {ILP} (2015)

\bibitem{KLOW18}
Konev, B., Lutz, C., Ozaki, A., Wolter, F.: Exact learning of lightweight
  description logic ontologies.
\newblock JMLR \textbf{18}(201), 1--63 (2018)

\bibitem{DBLP:conf/aaai/KonevOW16}
Konev, B., Ozaki, A., Wolter, F.: A model for learning description logic
  ontologies based on exact learning.
\newblock In: {AAAI}, pp. 1008--1015 (2016)

\bibitem{Guigues1986}
L., G.J., V., D.: Familles minimales d'implications informatives
  r{\'{e}}sultant d'un tableau de donn{\'{e}}es binaires.
\newblock Math{\'{e}}matiques et Sciences Humaines \textbf{95}, 5--18 (1986)

\bibitem{lehmann2009dl}
Lehmann, J.: {DL}-learner: learning concepts in description logics.
\newblock JMLR \textbf{10}, 2639--2642 (2009)

\bibitem{lehmann2010learning}
Lehmann, J.: Learning {O}{W}{L} class expressions, vol.~6.
\newblock IOS Press (2010)

\bibitem{lehmann2009ideal}
Lehmann, J., Haase, C.: Ideal downward refinement in the {E}{L} description
  logic.
\newblock In: {ILP}, pp. 73--87 (2009)

\bibitem{DBLP:journals/ml/LehmannH10}
Lehmann, J., Hitzler, P.: Concept learning in description logics using
  refinement operators.
\newblock Machine Learning \textbf{78}(1-2), 203--250 (2010)

\bibitem{lehmann2014perspectives}
Lehmann, J., V{\"o}lker, J.: Perspectives on Ontology Learning, vol.~18.
\newblock IOS Press (2014)

\bibitem{DBLP:journals/ijswis/Lisi11}
Lisi, F.A.: Al-quin: An onto-relational learning system for semantic web
  mining.
\newblock Int. J. Semantic Web Inf. Syst. \textbf{7}, 1--22 (2011)

\bibitem{DBLP:conf/aime/MaD13}
Ma, Y., Distel, F.: Learning formal definitions for {S}nomed {CT} from text.
\newblock In: {AIME}, pp. 73--77 (2013)

\bibitem{Maedche:2001:OLS:630317.630627}
Maedche, A., Staab, S.: Ontology learning for the semantic web.
\newblock IEEE Intelligent Systems \textbf{16}, 72--79 (2001)

\bibitem{McCulloch:1988:LCI:65669.104377}
McCulloch, W.S., Pitts, W.: A logical calculus of the ideas immanent in nervous
  activity.
\newblock In: Neurocomputing: Foundations of Research, pp. 15--27. MIT Press
  (1988)

\bibitem{muggleton1991inductive}
Muggleton, S.: Inductive logic programming.
\newblock New generation computing \textbf{8}(4), 295--318 (1991)

\bibitem{DBLP:conf/dlog/ObiedkovSZ19}
Obiedkov, S., Sertkaya, B., Zolotukhin, D.: Probably approximately correct
  completion of description logic knowledge bases.
\newblock In: {DL}, (2019)

\bibitem{DBLP:conf/rweb/Ozaki20}
Ozaki, A.: On the Complexity of Learning Description Logic Ontologies
\newblock In: {RW}, (2020)

\bibitem{DBLP:conf/aaai/OzakiPM20}
Ozaki, A., Persia, C., Mazzullo, A.: Learning Query Inseparable {ELH} Ontologies
\newblock In: {AAAI}, (2020)

\bibitem{Petrucci:2016:OLD:3092960.3092992}
Petrucci, G., Ghidini, C., Rospocher, M.: Ontology learning in the deep.
\newblock In: EKAW, pp. 480--495 (2016)


\bibitem{DBLP:conf/dlog/PorelloKRTGM19}
 Porello, D., 
                Kutz, O., 
                Righetti, G., 
                Troquard, N., 
                Galliani, P., 
                Masolo, C.: A Toothful of Concepts: Towards a Theory of Weighted Concept Combination.
\newblock In: DL  (2019)

\bibitem{DBLP:journals/corr/abs-2102-10689}
 Guimar{\~{a}}es, R.,
               Ozaki, A.,
               Persia, C., 
               Sertkaya, B.: Mining {EL} Bases with Adaptable Role Depth.
\newblock In: AAAI, (to appear) (2021)

\bibitem{Rudolph04exploringrelational}
Rudolph, S.: Exploring relational structures via {F}{L}{E}.
\newblock In: {ICCS}. Springer (2004)


\bibitem{DBLP:conf/semweb/SazonauS17}
Sazonau, V., Sattler, U.: Mining hypotheses from data in {OWL:} advanced
  evaluation and complete construction.
\newblock In: {ISWC}, pp. 577--593 (2017)

\bibitem{shalev2014understanding}
Shalev-Shwartz, S., Ben-David, S.: Understanding machine learning: From theory
  to algorithms.
\newblock Cambridge university press (2014)

\bibitem{DBLP:conf/rweb/0001GH18}
Stepanova, D., Gad{-}Elrab, M.H., Ho, V.T.: Rule induction and reasoning over
  knowledge graphs.
\newblock In: {RW}, pp. 142--172 (2018)

\bibitem{Valiant}
Valiant, L.G.: A theory of the learnable.
\newblock Commun. ACM \textbf{27}(11), 1134--1142 (1984)

\bibitem{DBLP:journals/ws/VolkerFS15}
V{\"{o}}lker, J., Fleischhacker, D., Stuckenschmidt, H.: Automatic acquisition
  of class disjointness.
\newblock Journal of Web Semantics \textbf{35}, 124--139 (2015)

\bibitem{volker2011statistical}
V{\"o}lker, J., Niepert, M.: Statistical schema induction.
\newblock In: The Semantic Web: Research and Applications, pp. 124--138.
  Springer (2011)

\bibitem{DBLP:conf/icml/WeissGY18}
Weiss, G., Goldberg, Y., Yahav, E.: Extracting automata from recurrent neural
  networks using queries and counterexamples.
\newblock In: Proceedings of the 35th International Conference on Machine
  Learning, {ICML} 2018, Stockholmsm{\"{a}}ssan, Stockholm, Sweden, July 10-15,
  2018, pp. 5244--5253 (2018)

\bibitem{DBLP:journals/corr/YangYHGD14a}
Yang, B., Yih, W., He, X., Gao, J., Deng, L.: Embedding entities and relations
  for learning and inference in knowledge bases.
\newblock In: {ICLR} (2015)

\bibitem{DBLP:journals/kbs/ZhuGPZXQ15}
Zhu, M., Gao, Z., Pan, J.Z., Zhao, Y., Xu, Y., Quan, Z.: Tbox learning from
  incomplete data by inference in belnet\({}^{\mbox{+}}\).
\newblock Knowl.-Based Syst. \textbf{75}, 30--40 (2015)

\end{thebibliography}

\end{document}